\documentclass[sigconf]{templates/ACMArt/acmart}

\usepackage{amsmath,amsfonts}
\usepackage{algorithmic}
\usepackage{graphicx}
\usepackage{textcomp}
\usepackage{xcolor}
\usepackage{makecell}
\usepackage[caption=false,font=normalsize,labelfont=sf,textfont=sf]{subfig}

\graphicspath{{figures}}

\AtBeginDocument{%
  \providecommand\BibTeX{{%
    \normalfont B\kern-0.5em{\scshape i\kern-0.25em b}\kern-0.8em\TeX}}}

\setcopyright{none}
\acmConference[XX]{XXXXX}{2023}{XXXX}

\begin{document}

\title{Assessing the Impact of a Supervised Classification Filter on Flow-based Hybrid Network Anomaly Detection}

\author{Dominik Macko}
\orcid{XXXX-XXXX-XXXX}
\affiliation{%
  \institution{Kempelen Institute of Intelligent Technologies}
  \city{Bratislava}
  \country{Slovakia}
}
\email{dominik.macko@kinit.sk}

\author{Patrik Goldschmidt}
\orcid{0000-0003-4872-1299}
\affiliation{%
    \small
    \institution{Faculty of Information Technology, Brno University of Technology}
    \city{Brno}
    \country{Czech Republic}
}
\affiliation{%
  \institution{Kempelen Institute of Intelligent Technologies}
  \city{Bratislava}
  \country{Slovakia}
}
\email{patrik.goldschmidt@kinit.sk}

\author{Peter Pi\v{s}tek}
\orcid{XXXX-XXXX-XXXX}
\affiliation{%
  \institution{Kempelen Institute of Intelligent Technologies}
  \city{Bratislava}
  \country{Slovakia}
}
\email{peter.pistek@kinit.sk}

\author{Daniela Chud\'{a}}
\orcid{XXXX-XXXX-XXXX}
\affiliation{%
  \institution{Kempelen Institute of Intelligent Technologies}
  \city{Bratislava}
  \country{Slovakia}
}
\email{daniela.chuda@kinit.sk}

\renewcommand{\shortauthors}{Macko, Goldschmidt, Pi\v{s}tek, and Chud\'{a}}

\begin{abstract}
Constant evolution and the emergence of new cyberattacks require the development of advanced techniques for defense. This paper aims to measure the impact of a supervised filter (classifier) in network anomaly detection. We perform our experiments by employing a hybrid anomaly detection approach in network flow data. For this purpose, we extended a state-of-the-art autoencoder-based anomaly detection method by prepending a binary classifier acting as a prefilter for the anomaly detector. The method was evaluated on the publicly available real-world dataset UGR'16. Our empirical results indicate that the hybrid approach does offer a higher detection rate of known attacks than a standalone anomaly detector while still retaining the ability to detect zero-day attacks. Employing a supervised binary prefilter has increased the AUC metric by over 11\%, detecting 30\% more attacks while keeping the number of false positives approximately the same.
\end{abstract}

\ccsdesc[500]{Security and privacy~Intrusion detection systems}
\ccsdesc[300]{Networks~Network security}
\ccsdesc[300]{Security and privacy~Network security}

\keywords{network intrusion detection, hybrid anomaly detection, supervised filter, network anomaly, network attack, machine learning}

\maketitle

\section{Introduction}
\label{sec:01}

Due to their enormous importance in today's world, computer networks are constantly under the watchful eyes of network administrators and various automated systems. They aim to monitor the network's behavior, communication, or other aspects to detect any unexpected behavior -- anomalies. Generally, we can classify anomalies based on their nature or casual aspects~\cite{chandola2009anomaly,fernandes2019comprehensive}. This fact implies that not every anomaly represents a network attack but can be caused by other events, such as a hardware failure or flash crowd, resulting from a successful marketing campaign. Nevertheless, network attacks represent an important subset of anomalies. For this purpose, intrusion detection systems (IDSs) are widely used.

Intrusion detection systems are automated defense and security systems for monitoring, detecting, and analyzing malicious activities within networks or end-host devices. IDSs can be categorized by several aspects, such as those discussed by Moustafa et al.~\cite{moustafa2019holistic}. In this paper, we focus on network-based IDSs (NIDSs), which aim to detect malicious behavior on computer networks. Furthermore, we also distinguish IDSs based on the two basic types of attack detection as \emph{misuse-based} (signature-based) for well-known attacks and \emph{anomaly-based} (behavior-based) to detect abnormal behavior.

Anomaly-based IDSs, compared to signature-based IDSs, can detect new attack types but with a significant increase in false positive ratios (mislabeled benign communication as attacks). Attack detection capabilities can be increased by combining these approaches in the so-called hybrid approach~\cite{fernandes2019comprehensive}. Such a combination then allows high accuracy and recall scores in detecting well-known attacks while the rest of the communication is examined for anomalies -- possible novel attacks. This approach enables automatic defense mechanisms for known attacks and decreases the number of anomalies required to be analyzed in depth.

Creating attack signatures or correctly identifying baseline traffic (i.e., normal, assumed benign) for IDS purposes requires expert knowledge and, thus, usually a high amount of manual effort. This issue can be tackled by machine learning techniques that can automate most of such tasks. Signatures of known attacks can be learned using supervised learning. It, however, requires labeled dataset. On the other hand, anomaly-based detection utilizes unsupervised or semi-supervised learning to detect traffic outliers or learn its baseline profile in an automated way. An advantage of these approaches comes in relaxing constraints on labels. Semi-supervised learning allows the combining of labeled and unlabeled data -- thus requiring fewer labels than supervised approaches. In contrast, unsupervised methods do not require any labeled data at all.

With the growth of network traffic and widespread encryption, a new problem has emerged. Nowadays, inspecting each packet in-depth separately and maintaining contextual information is rather challenging (see Section~\ref{ssec:discussion_netperf}). Instead, the NIDS community has begun experimenting with traffic metadata and communication statistics in the form of network flows~\cite{sperotto2010_flow_based_ids}. RFC 3954~\cite{rfc3954} defines a flow as a set of IP packets with common properties passing an observation point in the network during a certain time interval.

These common properties are typically specified as a 5-tuple based on source and destination IP addresses, L4 ports, and the L4 protocol type fields. Examples of collected communication statistics include the number of transmitted bytes, packets, or bytes per packet average. Therefore, instead of looking at the contents of every packet, a NIDS only processes communication statistics using much fewer computation resources. Newer flow collection protocols like NetFlow~\cite{rfc3954} v9 and IPFIX~\cite{rfc7011} provide new standardized flow statistics, allowing the detection of certain attack types more reliably. However, some attack types still become undetectable due to missing information from the packets' payload.

This paper performs an empirical evaluation of a supervised machine learning (ML) classifier (traffic prefilter) impact on anomaly detection in larger, Internet Service Provider (ISP)-sized networks. We achieve this by extending a SOTA anomaly detection method -- GEE by Nguyen et al.~\cite{nguyen2019gee} into a hybrid approach. Such an approach uses supervised learning for flow filtering by detecting known (learned) network attacks. The filtered traffic (i.e., assumed benign) is then processed by the unsupervised anomaly detector to detect deviations from the usual traffic patterns.

The intuition is that the hybrid approach will offer a higher detection rate of known attacks than standalone anomaly detection. Moreover, it should be able to detect zero-day (previously unknown) attacks without manually investigating excessive amounts of traffic, as the filtered attacks do not require further investigation. However, based on the current state of the research literature (using mostly smaller controlled environments for evaluation), it remains unknown whether such an existing hybrid approach is beneficial in larger-scale real networks. Based on these conditions, we frame our research questions as follows:

\begin{itemize}
    \item RQ1: \emph{When choosing a prefilter for hybrid anomaly detection in real large-scale networks, does a binary classifier outperforms a multi-class one?}
    \item RQ2: \emph{What is the impact of the classification-based prefilter on the detection and false alarm rates of anomaly detection in real large-scale networks?}
    \item RQ3: \emph{Does a hybrid anomaly detector perform better than binary classification regarding the detection of zero-day attacks in real large-scale networks?}
\end{itemize}

This work aims to obtain realistic findings by evaluating using the UGR'16 dataset~\cite{macia2018ugr}, containing real background traffic of an ISP. Such results are thus expected to reflect real-world scenarios better and make the discovered insights more likely to be utilizable for real-world scenarios. We hence define the following key paper contributions:

\begin{itemize}
\item measuring the impact of supervised traffic filtering onto anomaly detection by evaluating on a dataset containing traffic from a real large-scale network,
\item extending the SOTA autoencoder-based anomaly detector into a hybrid approach\footnote{Source code available at: \url{https://github.com/kinit-sk/hybrid-anomaly-detection}}, achieving higher detection rate of known attacks,
\item a more comprehensive evaluation of the SOTA anomaly detector GEE~\cite{nguyen2019gee} by exploring attack visibility phenomenon and possible temporal experimental bias in the used dataset.
\end{itemize}

The rest of the paper is organized as follows. The next section contains an overview of related work regarding hybrid approaches for anomaly detection in network flows. Section~\ref{sec:03} describes the employed hybrid approach for anomaly detection. In Section~\ref{sec:04}, we evaluate the model's performance and provide experimental results. Furthermore, we discuss method's performance and its specifics like detection delay, flow visibility, and temporal experimental bias in Section~\ref{sec:discussion}. Finally, the last section concludes the paper.

\section{Related Work}
\label{sec:02}

Network intrusion detection using machine learning techniques has been extensively researched in recent years~\cite{abdulganiyu2023_slr_nids,zhang2022_nids_comparative,ahmad2021_nids_systematic_study_ml}. However, most of the works focused on purely supervised approaches, so only a fraction aimed to detect anomalies~\cite{yang2022_slr_anids}, and even fewer of them by analyzing IP flow statistics~\cite{umer2017flow,jadidi2022automated}. A typical objective to approach such a problem would be to maximize a specific metric, such as accuracy or F-score. However, some research works focus on optimizing specific problems (e.g., minimizing false positives). Various methods to achieve such objectives can then be utilized.

Regarding supervised approaches, the community has employed a wide range of classifiers. For instance, Vinayakumar et al.~\cite{vinayakumar2019_dnn_ids} evaluated various classical ML approaches like Random Forest, Naive Bayes, or Support Vector Machine (SVM) and compared them to the proposed 5-layer Feed-forward Deep Neural Network. As a result, the neural network outperformed the classical ML models in most cases on eight IDS datasets, but some models, like Random Forest, achieved comparable results. Many others, primarily using deep learning~\cite{shone2018_dl_nids,yin2017_drnn_ids} or with various feature modifications~\cite{bagui2021_resampling_imbalanced_nids,alyaseen2017_multlevel_ids,muniyandi2012_nad_cascading}, were proposed. In these cases, it is important to emphasize that even if an IDS employs an unsupervised learner (e.g., for feature vectors clustering), it cannot be classified as hybrid because the unsupervised algorithm does not provide any anomaly-detection capabilities -- it influences the final output but the method still relies on supervised learning for attack detection. Recently, experiments with federated learning~\cite{tang2022_federated_nids} for NIDS while also trying to address other domain challenges, such as concept drift~\cite{andersini2021_insomnia}, were performed.

Although supervised approaches can learn known attacks with very high accuracy, they cannot detect attacks the method was not trained on. This behavior was investigated by various works, identifying a significant accuracy decrease when detecting new attacks~\cite{zoppi2023_ids_unknown_attacks,catillo2022_transferability_ml_ids,viegas2017toward}. Viegas et al.~\cite{viegas2017toward} observed that similar attacks could be detected with a slight increase in false positives (up to 3.95\%). Despite this fact, even such an increase in the false positive ratio significantly increases the absolute number of false alarms due to a huge traffic imbalance. Since all alarms need to be processed by other means (e.g., manually), the model might become unusable in practice as a result. This issue, known as a base-rate fallacy, was first examined by Axelsson in 1999~\cite{axelsson1999_base_rate_fallacy_ids} but is still highly relevant due to the ever-increasing amount of network traffic. The problem of high error cost is also identified by authors, who describe it as a potential obstacle for ML-based NIDS in real-world scenarios.

\begin{figure*}[t]
\vspace{-1.25em}
\centerline{
    \subfloat[\scriptsize Type 1 as in~\cite{khan2019novel,ji2016_multilevel_ids,deassis2014_7dim_flowanal,tombini2004_serial_anomaly_misuse_ids,barbara2001_adam}.]{
        \includegraphics[width=.255\linewidth]{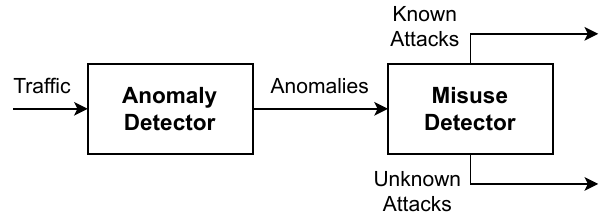}
    }
    \hspace{-0.5em}
    \subfloat[\scriptsize Type 2 as in~\cite{bangui2022hybrid,kim2014novel,elbasiony2013_hybrid_nids,zhang2008random}.]{
        \includegraphics[width=.255\linewidth]{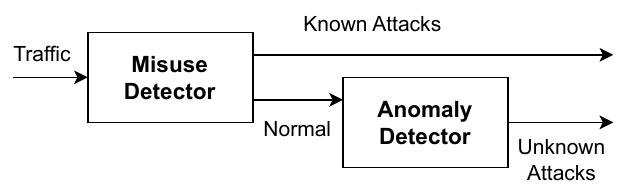}
    }
    \hspace{-0.5em}
    \subfloat[\scriptsize Type 3 as in~\cite{khraisat2020_hids_stacking,bhuyan2016_multistep_nad,depren2005_intelligent_ids}.]{
        \includegraphics[width=.255\linewidth]{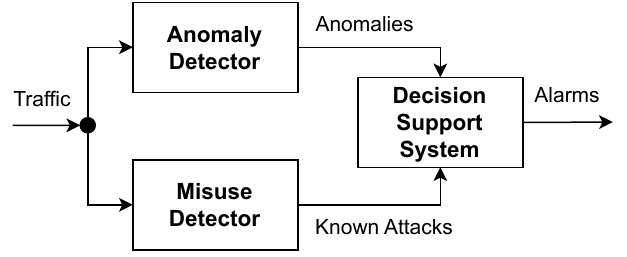}
    }
    \hspace{-0.5em}
    \subfloat[\scriptsize Type 4 as in~\cite{guo2016_twolevel_hybrid_ids}.]{
        \includegraphics[width=.255\linewidth]{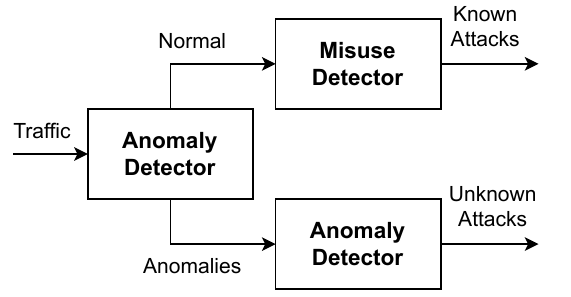}
    }
}
    \caption{Main types of hybrid NIDS designs.}
    \label{fig:nids_hybridized_designs}
\vspace{-1.25em}
\end{figure*}

To tackle the problem of unknown attacks, researchers have also been widely experimenting with anomaly-based IDS methods. The first works~\cite{lazarevic2003_nad_comparative} within this branch relied on simple statistical approaches, clustering, and algorithms based on distance, e.g., the Nearest Neighbors algorithm. Although they still achieve solid performance in specific use cases, the community has shifted towards more complex models for anomaly detection. Currently, models based on autoencoder architectures are the most popular for the task~\cite{yang2022_slr_anids,habeeb2022_nids_survey,ahmad2021_nids_systematic_study_ml}, although using Generative Adversarial Networks seems to be also raising into prominence~\cite{sabuhi2021_gans_nad}.

Various autoencoder (AE) architectures for network anomaly detection have been proposed recently. This includes variational AEs~\cite{nguyen2019gee}, AE ensembles~\cite{mirsky2018_kitsune}, stacked AEs~\cite{farahnakian2018_deep_ae_ids}, convolutional AEs~\cite{chen2018_autoencoder_nad}, and many more. Despite their efficiency in detecting unknown attacks, they still achieve relatively high false positive rates and are not as efficient in detecting known attacks as supervised solutions. Various works have tried to address this issue and make anomaly-based approaches more reliable. For instance, Li et al. ~\cite{li2020_aenc_ids_rf_fs} combined an autoencoder ensemble with an unsupervised Gaussian Mixture Models (GMM) (or K-means) along with feature selection using Random Forest and feature grouping. Such a model works in a fully unsupervised manner, increasing overall detection performance while still being relatively lightweight.
Similarly, Grill et al.~\cite{grill_reducing_2017} also successfully reduced false alarms by grouping the outputs of the anomaly detector into clusters.

In spite of numerous advances in purely supervised and purely unsupervised approaches, none of them could entirely cover up the benefits of one another. While unsupervised anomaly detection detects unknown attacks much better, it will hardly achieve the same performance for known attacks as supervised methods. Unsupervised approaches are known to have higher false positive and false negative rates, amplified by other challenges in the network intrusion detection domain~\cite{fernandes2019comprehensive,moustafa2019holistic}: missing or incomplete definition of normal behavior, severe concept drift, unavailability of reliable and standard dataset, and others.

The properties of both approaches thus imply that relying solely on one of them could lead the method to underperform in real-world deployments. This reasoning naturally leads to hybridized approaches and semi-supervised learning~\cite{vanengelen2020_survey_semisupervised}, able to combine benefits of both supervised and unsupervised methods~\cite{fernandes2019comprehensive}.

In general, four main types of hybrid NIDS designs have been presented in the literature~\cite{molinacoronado2020_survey_nids_kdd} (\figurename~\ref{fig:nids_hybridized_designs}). Type 1 scheme concatenates two detectors so that anomaly detection precedes misuse detection. Firstly, the anomaly detector produces suspicious items deviating from a normal profile. The misuse detector then searches for known patterns of attacks and false alarms. If one of such patterns is matched, the corresponding class is selected. If the misuse detector does not have high confidence for either case, the anomaly is classified as an unknown attack. Such an approach was presented by a famous IDS ADAM by Barbar\'{a} et al.~\cite{barbara2001_adam} in 2001 but was used in several other systems ever since~\cite{khan2019novel,ji2016_multilevel_ids,deassis2014_7dim_flowanal,tombini2004_serial_anomaly_misuse_ids}.

In the type 2 approach, as presented in~\cite{bangui2022hybrid,kim2014novel,elbasiony2013_hybrid_nids,zhang2008random}, the detectors are again serialized. This time, the misuse detector is placed first to detect known attacks. The remaining filtered traffic is passed to an anomaly detection module to detect possible unknown attacks. This way, known attacks are detected relatively reliably. At the same time, the anomaly detector can be less sensitive to deviations as in the case of type 1 (because some attacks are already detected), decreasing the false positive rate.

The type 3 approach utilizes misuse and anomaly detection modules in parallel. Both methods process all traffic, producing two sets of suspicious items. A decision support system (correlation component) is then used to determine whether to mark particular traffic as malicious or not. Such a system might be another ML method, weighted voting, or a simple threshold function. From the machine learning perspective, it can be understood as a stacking ensemble model. This approach was proposed by works such as~\cite{khraisat2020_hids_stacking,bhuyan2016_multistep_nad,depren2005_intelligent_ids}.

Finally, the type 4 hybrid system, proposed by Guo et al.~\cite{guo2016_twolevel_hybrid_ids}, employs two anomaly detectors and one misuse detector wired in a two-level hierarchy. In this setup, an anomaly detector is put first. Its non-anomalous traffic is then processed by a misuse module to identify undetected attacks from the first detector. On the other hand, anomalous traffic is forwarded to another more precise anomaly detector to confirm an attack. This effectively reduces the number of false positives but is rather computationally costly, as all input records have to be analyzed twice in a serial manner~\cite{molinacoronado2020_survey_nids_kdd}.

As outlined throughout this section, various approaches have been tried to maximize NIDS performance and reduce the number of false alarms. Nevertheless, despite the usage of hybrid systems being rather intuitive and generally believed to improve NIDS performance, there is a lack of rigorous analyses that would confirm this claim. Although some works included a comparison of their hybrid approach to a non-hybrid variant, the comparison was very limited~\cite{elbasiony2013_hybrid_nids,guo2016_twolevel_hybrid_ids,khraisat2020_hids_stacking} and none has focused on assessing the impact of a supervised element on the network anomaly detector.

One rather comprehensive analysis and the closest work to ours would be the one by Kim et al.~\cite{kim2014novel}. In this study, the authors evaluated an improvement of the 1-class SVM for anomaly detection by prepending a supervised C4.5 decision tree in an intrusion detection pipeline. Nevertheless, the analysis was performed on an NSL-KDD dataset, which is considered deprecated by now (see Section~\ref{ssec:04_data_selection}). As far as we are aware, no rigorous analysis for a modern large-scale dataset with an autoencoder-based model exists.

This paper aims to address the knowledge gap outlined in the previous paragraphs. For this purpose, we employ a type 2 hybrid IDS (with a misuse module first). In such a setup, we aim to evaluate the impact of a supervised classifier on unsupervised anomaly detection via ROC plots, KDE plots, and other metrics.

\section{Employed Approach for Hybrid Anomaly Detection}
\label{sec:03}

As outlined in previous sections, our work evaluates the impact of a supervised classification model used as a prefilter for anomaly detection. The evaluation is done using a hybrid anomaly detection method for network intrusion detection, similar to other published research works~\cite{bangui2022hybrid,kim2014novel,zhang2008random}.

\begin{figure}[!b]
\centerline{\includegraphics[width=0.95\linewidth,height=2cm,keepaspectratio,trim={0.9cm 0.9cm 0.5cm 0.5cm},clip]{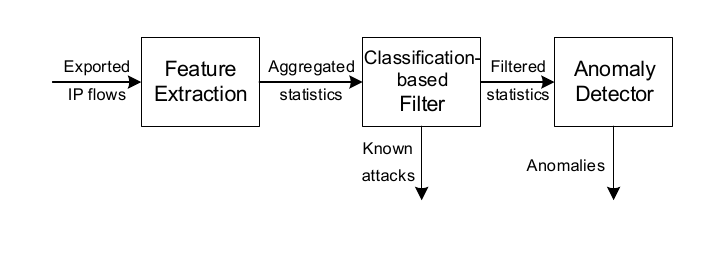}}
\caption{An overview of the used hybrid approach.}
\label{fig:proposal_arch}
\end{figure}

Since real-time per-packet processing within a cascade of ML models could be unsuitable for high-speed networks~\cite{suga2019_realtime_pkt_classification}, we aim to perform intrusion detection upon network IP flows. In such a setup, statistical information about communication is exported using Netflow v5/v9 or IPFIX protocols. The feature-extraction procedure (Section~\ref{ssec:ourappr_feature_extraction}) then converts flow statistics into aggregated time-window-based statistics per each source IP address. Such aggregated statistics are fed into the classification-based filter (Section~\ref{ssec:ourappr_classif_filter}), filtering out known attacks, while the rest of the traffic is analyzed in the anomaly detector (Section~\ref{ssec:ourappr_anomaly_detector}) (\figurename~\ref{fig:proposal_arch}).

\subsection{Feature Extraction}
\label{ssec:ourappr_feature_extraction}

In general, the utilized hybrid approach can be used with various feature extractors. However, to facilitate its validation and comparison with the original GEE framework, we utilize the same feature extraction procedure as in the former study by Nguyen et al.~\cite{nguyen2019gee}. This procedure aggregates the flow records into 3-minute time-window statistics for each source IP address, totaling 69 features:
\begin{itemize}
    \item 5 mean-based features,
    \item 5 standard deviation-based features,
    \item 5 entropy-based features,
    \item 54 port proportion-based features.
\end{itemize}
For details of the used aggregated features, please refer to Mun Hou's website~\cite{munhoublog}. In order for the statistics to be meaningfully representative, the authors decided to remove the samples (i.e., time windows of an IP) that aggregated less than ten flow records from the dataset. In order to achieve comparability of results, we kept this setting the same as in the original paper.

\subsection{Classification-based Filter}
\label{ssec:ourappr_classif_filter}

The classification-based filter component should reveal as many known attacks as possible. For this purpose, we propose to use a Random Forest classifier since it offers empirically similar performance as other algorithms (e.g., neural networks) while requiring a fraction of the training time~\cite{apruzzese2022_sok_unlabeled_data}. In our case of traffic filtering, both binary, as well as multi-class classifier variants could be utilized. The binary classifier can distinguish between malicious and benign traffic, while the multi-class classifier can also identify specific attack classes. Although offering some degree of explainability, we expect the multi-class classifier to perform worse than the binary one due to solving a potentially more complex problem. In this paper, we evaluate both classifiers to facilitate the selection of the most appropriate version (Section~\ref{ssec:04_results}).

\subsection{Anomaly Detector}
\label{ssec:ourappr_anomaly_detector}

As the anomaly-detector, we propose using a variational autoencoder (VAE)~\cite{kingma2013auto} model. The VAE is an unsupervised neural network model consisting of an encoder and a decoder part. The encoder is used to compute parameters for conditional distributions of the latent representation. The decoder is then used to generate samples from the latent representation. A reconstruction error (we use the mean squared error -- MSE) can then be used to measure how different the generated sample is in comparison to the original input feature vector. The reconstruction error should be low for the background traffic since the autoencoder will be trained on such traffic. Higher reconstruction error then represents a higher probability that the tested sample is an anomaly. We opted for this model due to its suitability for noisy data, like the network traffic in our case.

\section{Experimental Results}
\label{sec:04}

This section presents experimental results using the model described in Section~\ref{sec:03} upon the UGR'16 dataset~\cite{macia2018ugr}. As reliable public datasets are one of the more significant challenges connected with network intrusion and anomaly detection~\cite{ahmad2021_nids_systematic_study_ml,fernandes2019comprehensive,moustafa2019holistic}, we firstly reason about our dataset selection process in Section~\ref{ssec:04_data_selection}. Section~\ref{ssec:04_ugr_dataset} then briefly presents the chosen dataset and its specifics, while we further describe experiments methodology (Section~\ref{ssec:04_methodology}), setup (Section~\ref{ssec:04_exp_settings}), and finally, results (Section~\ref{ssec:04_results}).

\subsection{Dataset Selection}
\label{ssec:04_data_selection}

As reasoned in Introduction, we suggest our solution to utilize network flows rather than raw packet data. Therefore, we narrow our focus to datasets containing flow-based network traffic samples.

Despite their old age and very little relevance nowadays, datasets based on DARPA98~\cite{lippmann20001999} data, namely KDD99~\cite{bay2000uci} and NSL-KDD~\cite{tavallaee2009_nslkdd} are still used in more than a half of the published research articles nowadays~\cite{hindy2020_network_threats_taxonomy,habeeb2022_nids_survey}. Based on various criticisms, such as by Al et al.~\cite{al2018kdd} and Siddique et al.~\cite{siddique2019kdd}, these datasets are considered obsolete, not corresponding to reality and thus are strongly recommended not to be used for NIDS benchmarking.

Other well-known datasets suitable for our case, like Sperotto~\cite{sperotto2009labeled}, CTU-13~\cite{garcia2014empirical}, and UNSW-NB15~\cite{moustafa2015unsw}, are also rather older now, so they usually contain different network profiles compared to the contemporary ones. Datasets CIC-IDS2017 and especially CSE-CIC-IDS2018~\cite{sharafaldin2018toward} looked particularly encouraging due to their size, infrastructure, and novelty, but they do not contain any data from real-world environments. Moreover, Engelen et al.~\cite{engelen2021troubleshooting} discovered multiple issues within the tool (CICFlowMeter) used to generate CIC-IDS2017 data. Since the same tool used in CIC-IDS2017 for flow creation was also used in CICIDS2018, the same problems likely remain. Although a corrected CIC-IDS2017 version was also released, the dataset contains several features that are not part of NetFlow v5/v9 or IPFIX standards, and thus not correspond to realistic flow-based feature vectors available in-the-wild.

Although suitable for our purposes of determining supervised filtration efficiency, we filtered out datasets focused on a specific network types, like InSDN~\cite{elsayed2020_insdn} for software-defined networks or MedBIoT~\cite{guerramanzanes2020_medbiot} or TonIoT~\cite{moustafa2021_toniot} for Internet of Things networks. Hence, we further narrow our scope of experiments on flow-based NIDS in general computer networks.

Finally, we decided to use UGR'16~\cite{macia2018ugr} by Marci\'{a}-Fern\'{a}ndez et al. due to its real-world nature and relatively good establishment among the community, as several dozens of works have already used it. Another significant reason was that the original GEE framework, we utilized for feature extraction and the anomaly detection module, used this dataset as well, making comparisons to the original work without the prefilter simpler. Nevertheless, other relatively current and promising datasets such as LITNET-2020\cite{damasevicius2020_litnet2020}, CUPID~\cite{lawrence2022_cupid}, or Simargl2022~\cite{komisarek2022_simargl2022} meet (most of) our criteria as well and can be an interesting way to test models' performance in future research. Please note that this section is not a comprehensive review and critique of existing NIDS datasets, and many other, relevant for other specific use-cases exist.

\subsection{UGR'16 Dataset}
\label{ssec:04_ugr_dataset}

As discussed in the previous subsection, we evaluate using the UGR'16 dataset~\cite{macia2018ugr}. The dataset contains real-world traffic spanning over six months captured from a Tier 3 Internet Service Provider (ISP) network. In total, the dataset contains 16.7 billion unidirectional flow records collected using NetFlow v9. The traffic was spiced by synthetically generated attacks such as Denial of Service (DoS) or botnet traffic. In addition, real background traffic was screened by three anomaly detectors and confirmed anomalies appropriately labeled.

The dataset itself contains nine classes in total. The class \emph{dos} was performed as a low-rate TCP SYN flood with various attack profiles on the port 80. Therefore, it is mixed with background traffic from different sources. Port scanning was simulated with SYN packets. According to its variant, \emph{scan11} labels represent a one-to-one scanning, and \emph{scan44} corresponds to a situation when four attackers initiate a scan to four victims at once. Botnet traffic, labeled \emph{nerisbotnet}, marks the execution of malware known as Neris. As it would be infeasible to infect a host on a real network, the UGR'16 authors obtained publicly-available botnet traffic, converted it to flows, and inserted it into their dataset by modifying the flow timestamps and IP addresses to be coherent with other data. The inserted records might have different traffic characteristics due to its origin in a different network environment, yet, it is still valuable that real botnet traffic is present in the dataset.

Traffic with other labels was captured directly from the Tier 3 ISP and labeled with various techniques. Four public denylists for malware, ransomware, spam, etc., were used to label flows corresponding to IP addresses present in a used denylists with the \emph{blacklist} tag (original tag name in the UGR'16 dataset). Labels \emph{anomaly-spam}, \emph{anomaly-udpscan}, and \emph{anomaly-sshscan} were produced by using three different anomaly detectors. The detected anomalies were then manually investigated by the authors for confirmation. The rest of captured traffic from the ISP got the \emph{background} tag. Despite the dataset authors' best efforts, it must be noted that one cannot be sure that it does not contain any more (unidentified) attacks. Also, the flow records of IP addresses included in public denylists do not necessarily mean that attacks are recorded in the corresponding flows.

\subsubsection{UGR'16 Subset}

In order to compare with the original GEE framework~\cite{nguyen2019gee}, we use the same subset of the UGR'16 dataset as the original study. This subset consists of two days for training (March 19, July 30) and three days for testing (March 18, March 20, July 31). In such a setup, we are aware of temporal training consistency violation (as defined by Pendlebury et al.~\cite{pendlebury2019_tesseract}) and, thus, the existence of a potential temporal bias that may skew the experimental results. Nevertheless, to retain comparability with the original study, we continue our evaluation using this dataset subset and discuss the temporal bias issue in Section~\ref{ssec:discussion_tempbias}.

Employed GEE-based feature extraction aggregates the flow records from the UGR'16 dataset into time-window-based statistics for each source IP address. As already mentioned, samples aggregating less than 10 flows were removed from the dataset. Therefore, the resulting number of window-based samples is much smaller than the declared number of UGR'16 flow records. The counts of preprocessed samples per every label are provided in Table~\ref{tab:01}.

\begin{table}[!b]
\caption{UGR'16 Train and Test Subsets Labels Overview After Preprocessing}
\begin{center}
\begin{tabular}{|c|c|c|}
    \hline
    \textbf{Class label} & \textbf{Train} & \textbf{Test} \\
    \hline
    background & 1,994,486 & 2,591,933 \\
    blacklist & 4,470 & 5,726 \\
    nerisbotnet & 983 & 985 \\
    anomaly-spam & 384 & 822 \\
    dos & 67 & 72 \\
    scan44 & 30 & 32 \\
    scan11 & 8 & 7 \\
    \hline
    \textbf{Total} & 2,000,428 & 2,599,577 \\
    \hline
\end{tabular}
\label{tab:01}
\end{center}
\end{table}

\subsection{Methodology}
\label{ssec:04_methodology}

As briefly outlined in Section~\ref{ssec:ourappr_classif_filter}, we can use either a binary or multi-class classifier for traffic filtering purposes. A multi-class classifier offers a better explainability of the classified attacks. In contrast, binary classification is expected to offer better performance since it must distinguish only between an attack and background traffic. Our first set of experiments thus seeks to verify this expectation in large-scale environments, formulated as the first research question: \textit{RQ1: When choosing a prefilter for hybrid anomaly detection in real large-scale networks, does a binary classifier outperform a multi-class one?} In order to properly compare them, we modified multi-class predictions into a binary form (all attack classes represent the class $1$, and the background represents the class $0$).

Our employed hybrid approach for anomaly detection combines a classifier with an autoencoder. We illustrate its benefits by comparing it with the original GEE VAE implementation~\cite{munhoublog,munhougithub} (a sole unsupervised anomaly detector), used as the anomaly detector of the employed hybrid approach. The autoencoder was trained using only the background traffic of the discussed train set. The same pre-trained model was run using the GEE test set with and without the previous binary classifier's filtering.

In the second research question: \textit{RQ2: What is the impact of the classification-based prefilter on the detection and false alarm rates of anomaly detection in real large-scale networks?}, we seek to answer whether the supervised prefilter truly helps in anomaly detection in real larger networks. The filtering was done so that the anomaly detector's reconstruction error (MSE) was considered only for traffic classified as a background by the binary classifier. For comparison, the reconstruction error of the anomaly detector was set to $1.0$ for samples classified as attacks by the binary classifier (i.e., filtered).

Classification is known to provide better results in the detection of learned attacks. However, it should perform poorly in detecting unknown (zero-day) attacks. Therefore, the third set of experiments compares the adopted hybrid approach to the standalone binary classifier in larger networks: \textit{RQ3: Does a hybrid anomaly detector perform better than
binary classification regarding the detection of zero-day attacks
in real large-scale networks?} For this purpose, we have used novelty tests, intentionally omitting some attack classes (each time different) during the classifier training (referred to as an open world scenario tests~\cite{apruzzese2023_sok_pragmatic}).

We have used the following performance metrics: the Area under the curve of Receiver Operating Characteristic (AUC), recall for class $1$ (malicious), the number of true positives, and the number of false positives. 

\subsection{Experiments Settings}
\label{ssec:04_exp_settings}

For classification-based filtering, we used the scikit-learn~\cite{scikit-learn} implementation of Random Forest with default parameters. Some of its key default parameters include 100 trees and the Gini impurity as a measure of split quality.

\begin{figure}[b]
\centerline{\includegraphics[width=\linewidth]{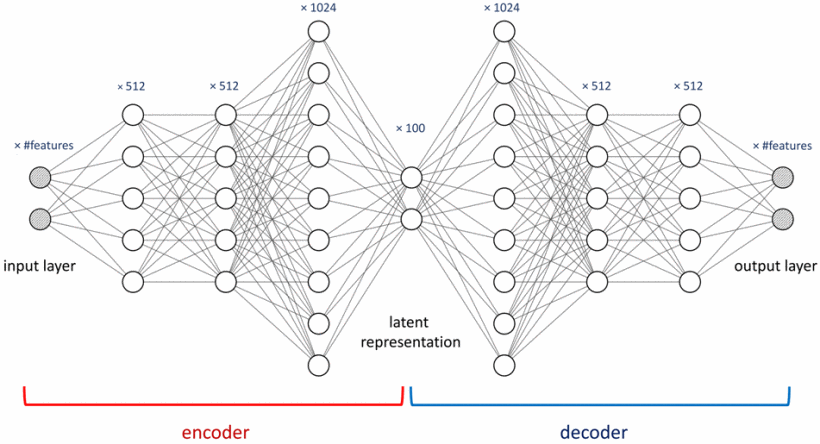}}
\caption{The variational-autoencoder architecture~\cite{nguyen2019gee}.}
\label{fig:varenc_arch}
\vspace{-5mm}
\end{figure}

Regarding the neural network architecture, we have used the same architecture as the original anomaly detector~\cite{nguyen2019gee} (\figurename~\ref{fig:varenc_arch}), against which we have compared our modified hybrid approach. Autoencoder is designed with an input layer (69 neurons), three encoder hidden layers (512, 512, and 1024 neurons), three decoder hidden layers (1024, 512, and 512 neurons), and an output layer (69 neurons).  All of these are linear layers. The latent representation has a size of 100. As an activation function, all intermediate layers employ the rectified linear unit (ReLU), while the sigmoid is present at the output. Its training was done by the Adam optimizer with the learning rate of $0.001$ and the $0.01$ weight decay.

For the autoencoder's training, only background traffic from the train dataset was used. The classifier was trained on a balanced train subset. As we aimed to test both binary and multi-class classifiers, two kinds of balanced training subsets were created.

Binary training was performed by selecting background and attack (anomaly) traffic in the ratio of $1:1$. More specifically, random undersampling to 6000 samples was used for background, and random undersampling and oversampling to 1000 samples were used for attack classes. Thus, 12000 samples were included in this balanced subset. When some attack class was omitted from the training for novelty test experiments, 1000 background samples were also randomly removed to keep the balancing ratio intact. During the multi-class training, each class was randomly undersampled or oversampled to 1000 samples, totaling 7000 samples.

Note that we have not performed hyperparameter tuning for both supervised and non-supervised models, as it brings little relevance to the purpose of this paper to answer our research questions. 

As elaborated on by the GEE method's original authors~\cite{nguyen2019gee}, the \emph{blacklist} attack class of the UGR'16 dataset is indistinguishable from the background based on source IP addresses' behavior (statistics). Therefore, we have omitted this class from further experiments by not including it in the train and test sets.

\subsection{Results}
\label{ssec:04_results}

For the purpose of traffic filtering, we have compared the binary and multi-class classifiers in the first set of experiments. The superior performance of a binary classification has already been shown by other works, e.g., empirically by Acharya et al.~\cite{acharya2021_heterogenous_ensemble}, but also in a statistically significant manner by Apruzzese et al.~\cite{apruzzese2023_sok_pragmatic}. However, it remains unclear whether the binary classifier also performs better in a larger network with real background traffic. Our results are shown in Table~\ref{tab:02}).

\begin{table}[b]
\caption{Binary and Multi-class Classifiers Comparison}
\begin{center}
\begin{tabular}{|c|c|c|c|c|}
\hline
\textbf{Classifier} & \textbf{AUC} & \textbf{Recall(1)} & \textbf{TP} & \textbf{FP} \\
\hline
Binary (train) & 0.9975 & 0.9966 & 1467 & 3263 \\
Binary (test) & 0.9837 & 0.9692 & 1859 & 4948 \\
Multi-class (train) & 0.9961 & 0.9973 & 1468 & 10264 \\
Multi-class (test) & 0.9560 & 0.9176 & 1760 & 14388 \\
\hline
\textbf{\scriptsize Binary vs multi-class (train)} & +0.14\% & -0.07\% & -1 & -7001 \\
\textbf{\scriptsize Binary vs multi-class (test)} & +2.76\% & +5.16\% & +99 & -9440 \\
\hline
\multicolumn{5}{l}{\scriptsize $AUC$ represents the area under the curve of Receiver Operating Characteristic.}\\
\multicolumn{5}{l}{\scriptsize $Recall(1)$ represents recall metric for class $1$.}\\
\multicolumn{5}{l}{\scriptsize $TP$ is the number of true positives (i.e., correctly identified attacks).}\\
\multicolumn{5}{l}{\scriptsize $FP$ is the number of false positives (i.e., background traffic predicted as an attack).}
\end{tabular}
\label{tab:02}
\end{center}
\vspace{-5mm}
\end{table}

As we can see, the binary classifier provides a higher AUC metric. More specifically, it has approximately the same number of true positives with significantly fewer false positives (over 60\%). Therefore, we find it a better choice, even when explainability is lost (we do not know which specific attack class a sample belongs to). Nevertheless, explainability ability could be introduced by cascading binary and multi-class classifiers as in~\cite{apruzzese2023_sok_pragmatic}, while retaining the same performance metrics.

Based on the results, our experiments confirm the conclusions of other works that the binary classification indeed outperforms multi-class even in realistic settings such as those provided by the UGR'16 dataset. This thus answers our first research question.

\begin{figure*}[!t]
\vspace{-1.25em}
\centerline{
    \subfloat[\scriptsize The original (unsupervised) anomaly detector.]{
        \includegraphics[width=.275\linewidth]{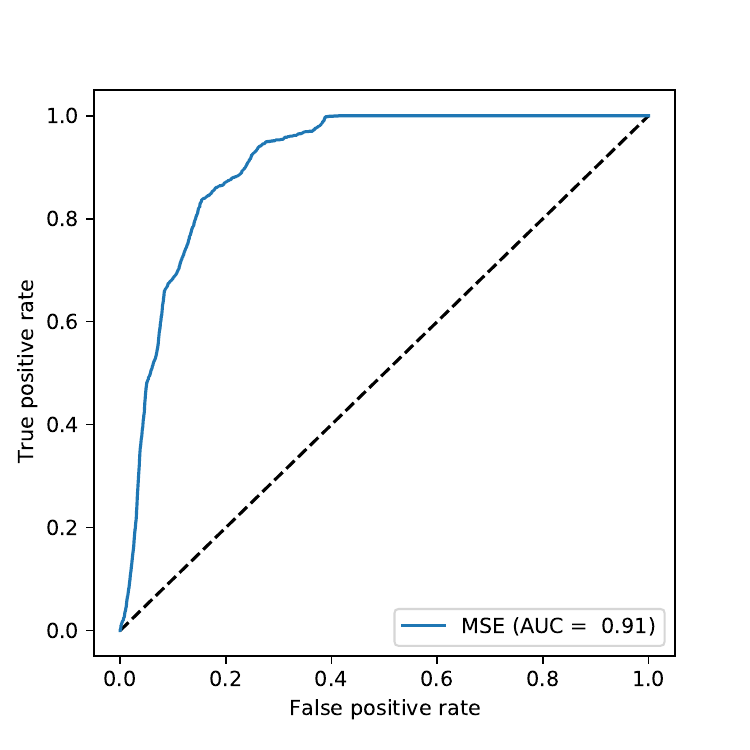}
        \hspace*{-1.2em}
        \includegraphics[width=.275\linewidth]{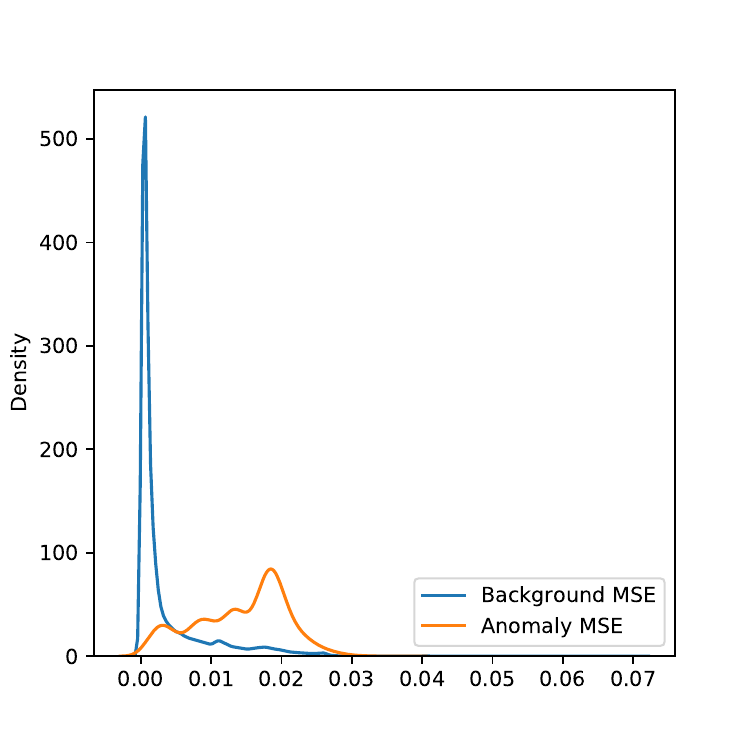}
    }
    \hspace*{-1.5em}
    \subfloat[\scriptsize The employed (hybrid) anomaly detector.]{
        \includegraphics[width=.275\linewidth]{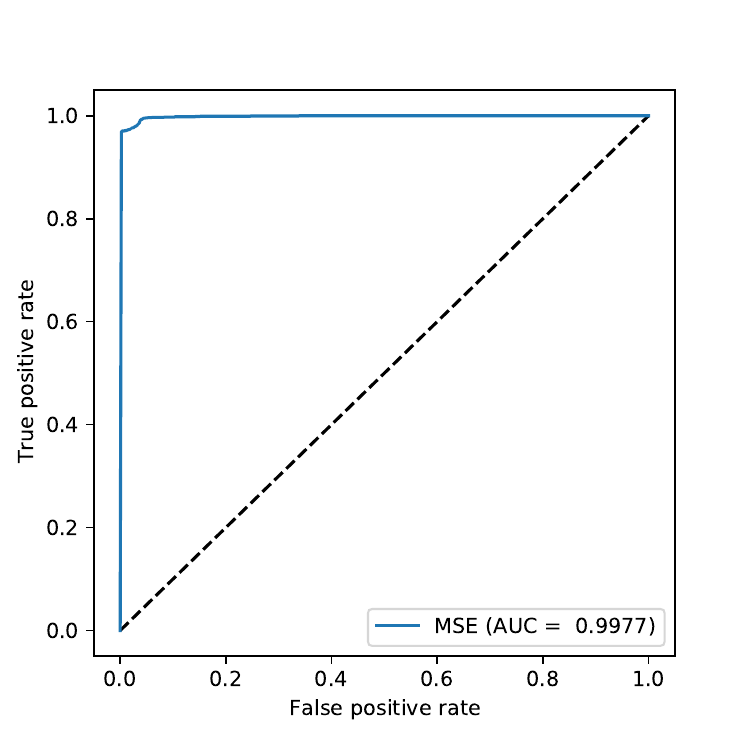}
        \hspace*{-1.2em}
        \includegraphics[width=.275\linewidth]{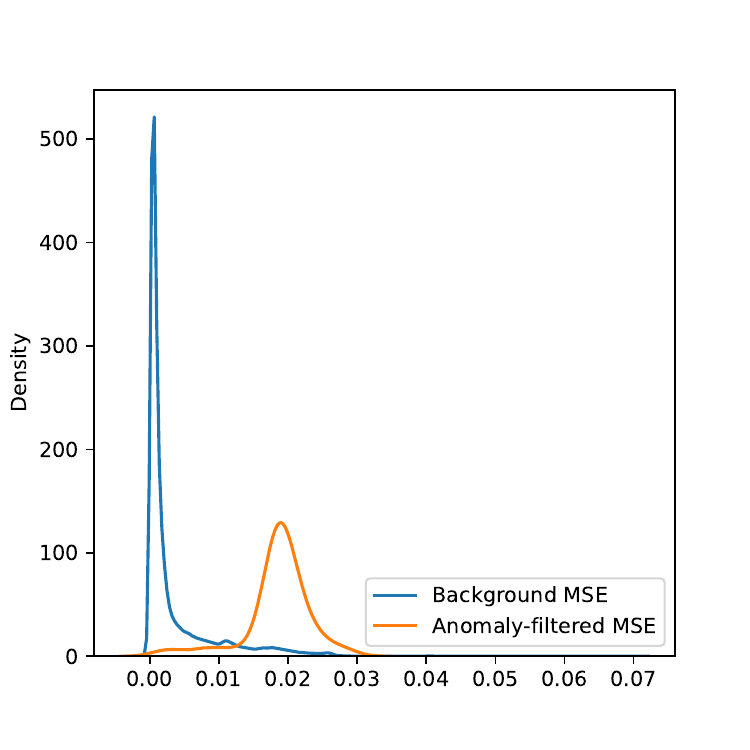}
    }
}
\caption{Comparison of the original unsupervised (a) and the employed (b) anomaly detection: the first (left) in each pair represents the ROC curve using reconstruction error, while the second represents the KDE plot of the reconstruction errors.}

\label{fig:roc_kde_comp}
\end{figure*}

The second set of experiments was focused on comparing the hybrid anomaly detector with the original one (unsupervised only).  \figurename~\ref{fig:roc_kde_comp} illustrates receiver operating characteristic (ROC) curve and kernel density estimation (KDE) plot of the reconstruction errors. Anomalies with the reconstruction error of $1.0$ are not included in the KDE plot.

As seen in the KDE plot of the original detector (top right), the threshold for distinguishing anomalies and background traffic is challenging to select. On the other hand, the threshold for the modified detector is rather clear, somewhere around $0.01$ of the reconstruction error. Note that the threshold should not be chosen based on the test dataset, yet we wanted to illustrate how was the problem of traffic reconstruction simplified by employing a supervised prefilter. The original detector's KDE plot is very similar even for the train dataset -- anomalous and background traffic still not being clearly distinguishable, while the same problem does not remain for the modified detector's case. These facts imply that the binary classifier has successfully filtered a significant part of the attacks, simplifying the anomaly threshold selection and thus increasing the number of correct anomaly detections.

For further evaluation, we have selected the reconstruction-error threshold ($\tau$) using the formula:
\begin{equation}
\tau = \mathcal{L}_{\mu} + \mathcal{L}_{\delta}
\label{eq:01}
\end{equation}
where $\mathcal{L}_{\mu}$ is mean and $\mathcal{L}_{\delta}$ is standard deviation of reconstruction error (loss) values for the training dataset. We have tried multiple coefficients with the $\mathcal{L}_{\delta}$ component, ranging from $\frac{1}{3}\mathcal{L}_{\delta}$ to $3\mathcal{L}_{\delta}$ . As found out, the threshold formula from Eq.~\ref{eq:01} (the coefficient of~1) achieved the highest F1 score for the attack class (as a measure to keep the number of true positives as high as possible and at the same time the number of false positives as low as possible).

The comparison of the results, using the selected reconstruction-error threshold of $0.008818$, is provided in Table~\ref{tab:03}. We can see that the modified anomaly detector can correctly detect a higher number of attacks (\textasciitilde30\% more) than the original anomaly detector, while the increased number of false positives is negligible (0.16\%). Overall, the AUC metric increased by 11.56\%. Note that this AUC value is different from Figure~\ref{fig:roc_kde_comp}, as in this case, we calculated it from binary decisions (i.e., background/malicious) based on the chosen threshold. In contrast, AUC in Figure~\ref{fig:roc_kde_comp} was calculated using the MSE score without any thresholding.

\begin{table}[!b]
\caption{Comparison of the modified hybrid to unsupervised anomaly detection. Recall(1) represents recall metric for the class 1 (attacks).}
\begin{center}
\begin{tabular}{|c|c|c|c|c|}
\hline
\textbf{Anomaly detector} & \textbf{AUC} & \textbf{Recall(1)} & \textbf{TP} & \textbf{FP} \\
\hline
Original & 0.8174 & 0.7664 & 1470 & 341274 \\
Modified & 0.9330 & 0.9979 & 1914 & 341827 \\
\hline
\textbf{Modified vs Original} & +11.56\% & +23.15\% & +444 & +553 \\
\hline
\end{tabular}
\label{tab:03}
\end{center}
\vspace{-5mm}
\end{table}

The obtained results have already clearly shown the benefits of the hybrid approach with the supervised element against original (unsupervised) anomaly detection. The above elaboration thus answers our second research question, and we can conclude that using a supervised prefilter before an anomaly detection module can significantly improve the overall anomaly detection capabilities, while having negligible effect on the false alarm rate.

However, we may notice that the sole binary classifier performs even better. When we compare the performance metrics of the modified anomaly detector in Table~\ref{tab:03} to those of the binary classifier in Table~\ref{tab:02}, we observe AUC decrease by approx. 5\%. However, the modified detector revealed about 3\% more attacks, though the number of false positives has significantly increased. Nevertheless, such results are expected since the detection of learned attacks is known to perform better than the detection of anomalies. Still, the highest advantage of anomaly detection is the ability to detect unknown (zero-day) attacks. It is then up to the organization's security policy whether to prefer better attack detection at the cost of more false alarms or instead choose better reliability. Nevertheless, it must be noted that all possible attacks cannot be learned, and so supervised approaches will miss a non-negligible portion of attacks by design.

To illustrate the advantage of the employed hybrid approach (the ability of zero-day attack detection) compared to the standalone binary classifier, we have performed experiments using novelty tests. The novelty tests involve omitting a specific attack class from classifier training to imitate a new attack type in the test dataset. We have tested the modified hybrid anomaly detector and the standalone binary classifier using all attacks (besides the \emph{blacklist} due to indistinguishability). The results are provided in Table~\ref{tab:04}. The first column refers to classes omitted during the classifier's training. The rest of the columns contain values for the AUC and recall metrics (as in the previous tables).

\begin{table}[!b]
\caption{Comparison of the proposal to the binary classifier using novelty tests. Class \emph{blacklist} was omitted in all cases.}
\begin{center}
\begin{tabular}{|l|c|c|c|c|}
\hline
\textbf{Omitted Classifier} & \multicolumn{2}{|c|}{\textbf{Proposal}} & \multicolumn{2}{|c|}{\textbf{Binary Classifier}}\\
\cline{2-5}
\textbf{Training Classes} & \textbf{\textit{AUC}} & \textbf{\textit{Recall(1)}} & \textbf{\textit{AUC}} & \textbf{\textit{Recall(1)}} \\
\hline
dos & \textcolor{black}{0.9163} & \textbf{0.9645} & \textbf{0.9460} & \textcolor{black}{0.8942} \\
nerisbotnet & \textbf{0.8434} & \textbf{0.8186} & \textcolor{black}{0.7082} & \textcolor{black}{0.4176} \\
spam & \textbf{0.9328} & \textbf{0.9974} & \textcolor{black}{0.7833} & \textcolor{black}{0.5667} \\
scan11 & \textcolor{black}{0.9324} & \textbf{0.9969} & \textbf{0.9828} & \textcolor{black}{0.9677} \\
scan44 & \textcolor{black}{0.9319} & \textbf{0.9958} & \textbf{0.9656} & \textcolor{black}{0.9333} \\
scan & \textcolor{black}{0.9234} & \textbf{0.9791} & \textbf{0.9826} & \textcolor{black}{0.9682} \\
\hline
\end{tabular}
\label{tab:04}
\end{center}
\vspace{-5mm}
\end{table}

Due to attack-class imbalance in the test dataset and to better see performance for individual attack classes, we have also tested the models using only background and the selected attack class (the one omitted from the training). The results are provided in Table~\ref{tab:05}.

\begin{table}[!b]
\caption{Comparison of the hybrid detector to binary classifier using novelty tests evaluated on background and omitted attack class (\emph{blacklist} was always omitted).}
\begin{center}
\begin{tabular}{|l|c|c|c|c|}
\hline
\textbf{Omitted Classifier} & \multicolumn{2}{|c|}{\textbf{Proposal}} & \multicolumn{2}{|c|}{\textbf{Binary Classifier}}\\
\cline{2-5}
\textbf{Training Classes} & \textbf{\textit{AUC}} & \textbf{\textit{Recall(1)}} & \textbf{\textit{AUC}} & \textbf{\textit{Recall(1)}} \\
\hline
dos & \textcolor{black}{0.4896} & \textbf{0.1111} & \textbf{0.4990} & \textcolor{black}{0.0000} \\
nerisbotnet &  \textbf{0.7585} & \textbf{0.6487} & \textcolor{black}{0.4993} & \textcolor{black}{0.0000} \\
spam & \textbf{0.9323} & \textbf{0.9964} & \textcolor{black}{0.4999} & \textcolor{black}{0.0000} \\
scan11 & \textcolor{black}{0.9340} & \textbf{1.0000} & \textbf{0.9990} & \textbf{1.0000} \\
scan44 & \textcolor{black}{0.8871} & \textbf{0.9062} & \textbf{0.9520} & \textbf{0.9062} \\
scan & \textcolor{black}{0.4851} & \textbf{0.1026} & \textbf{0.4985} & \textcolor{black}{0.0000} \\
\hline
\end{tabular}
\label{tab:05}
\end{center}
\vspace{-5mm}
\end{table}

As we can see, the hybrid detector offers a higher recall for all attack classes, answering our third research question. The classifier cannot detect attacks, which it was not trained on (the recall of 0\%). However, due to their similar nature, this supposition is broken for attack classes \textit{scan11} and \textit{scan44}. When both of them are omitted from the training (the last row), the classifier cannot detect them at all. The hybrid detector could always detect at least 10\% of the attacks. When \textit{nerisbotnet} or \textit{anomaly-spam} attack classes were omitted from training, the hybrid detector provided even higher AUC than the sole classifier.

\section{Discussion}
\label{sec:discussion}

The results presented in this paper confirm the superiority of the employed hybrid approach with regard to the original unsupervised one. This demonstrates the benefits of a supervised classifier element (prefilter) employed within the anomaly detection pipeline. Nevertheless, several considerations have to be made about the method and its functionality, discussed in the following subsections.

This section will further discuss the following topics: 
\begin{itemize}
 \item network performance (Section~\ref{ssec:discussion_netperf}),
 \item detection delay (Section~\ref{ssec:discussion_detection_delay}),
 \item flow visibity (Section~\ref{ssec:discussion_flowvis}),
 \item temporal experimental bias (Section~\ref{ssec:discussion_tempbias}),
 \item comparison with other approaches (Section~\ref{ssec:discussion_comparison}).
\end{itemize}

\subsection{Network Performance}
\label{ssec:discussion_netperf}

Deep learning algorithms tend to be rather computationally expensive and often struggle to keep up with incoming network traffic in real-time. For instance, the authors of the SOTA ENIDrift NIDS~\cite{enidrift} report their approach to reach the processing of 49 packets/s, while another well-known online NIDS Kitsune~\cite{mirsky2018_kitsune} reaches only 7 packets/s on average. Nevertheless, packet rates in existing datasets captured on larger networks are significantly higher, reaching 382 packets/s in CIC-IDS2017~\cite{sharafaldin2018toward} and over 37,000 packets/s in UGR'16~\cite{macia2018ugr} (our estimate). As already outlined throughout this document, we have partially mitigated this issue by using flow records rather than per-packet data. Nevertheless, we still aim to measure the throughput of the method in order to determine its ability to keep pace with the network activity.

Based on the number of flows in each fully captured day, we approximate the throughput of the UGR'16 network to be 1,273 flows/s (given \textasciitilde110M flows/day). Ideally, the employed hybrid approach must achieve higher speeds to cover up traffic peaks. Therefore, we have measured the method's performance using a single machine with two AMD EPYC 7302 16-Core CPUs and 256 GB RAM.

On average, the feature extraction performs time-window aggregation at over 19,000 flows/s, the model input preparation (i.e., value normalization), then at 118,000 aggregated samples/s. Regarding detection models, the Random Forest classifier runs at 150,000 aggregated samples/s, while the variational autoencoder (i.e., computation of reconstruction errors) over 8,000 aggregated samples/s.

Statistically, an average time-window aggregated sample contains 25 flows. Therefore, when run as a whole serial pipeline, the method can process approximately 17,000 flows/s. This indicates that the discussed hybrid anomaly detector would be useful for real-time processing even for networks of higher speeds than the network in the UGR'16 dataset. Note that this performance was achieved purely by CPU computation, and further speed-ups could be achieved by GPU acceleration and horizontal scaling.

\subsection{Detection Delay}
\label{ssec:discussion_detection_delay}

When discussing the method's detection delays, two factors must be considered: 1) the time-window aggregation mechanism and 2) flow export delays.

As discussed, the data preprocessing mechanism aggregates incoming flow records into 3-minute time windows. Anomaly detection is thus delayed by 3 minutes. The delay can indeed be tweaked, but decreasing the time window length might negatively impact the model's detection capabilities due to less statistical information collected during a shorter time frame.

Since the method relies on network flows, specifics of the mechanisms creating them also need to be considered. Network flows are typically created with NetFlow/IPFIX-enabled devices. Flows are created on an exporter, and after the flow is finished or a timeout occurs, it gets transferred to a collector, possibly collecting data from multiple exporters. Such a collector device can then provide flows for further analysis or anomaly detection.

The detection is hence further delayed by the time needed to exchange data between the exporter and the collector, as well as query the collector for new data. Most exporters are configured to export flows in bulks, so even if a flow is finished, additional time is needed until it arrives at the collector. A flow is considered finished when a TCP FIN flag is seen. Attacks such as DoS typically do not carry FIN segments, so the timeout period (typically 30 to 120 seconds) must also be considered. To sum up, the flow mechanism itself delays attack detection from several seconds up to a few minutes based on its configuration.

\subsection{Flow Visibility}
\label{ssec:discussion_flowvis}

Another specific property of the method is that samples for anomaly detection are not created if there are 10 or fewer flows for a given source IP address in a 3-minute aggregation window (see~\cite{nguyen2019gee}). For this reason, some attacks might not be included in the evaluation.

To examine a difference, we have analyzed the remained flow visibility after feature extraction. For this purpose, we disabled the minimal number of flows in a time window constraint (preset by the authors of the original anomaly detector~\cite{nguyen2019gee}). Afterward, we calculated the percentage of the total number of flows aggregated to time-window-based samples from less than 10 flows with respect to all samples. The results are provided in the left part of Table~\ref{tab:06} for each class. As the results show, the number of omitted attacks is rather insignificant (under 1\%). Although this claim is invalid for the \emph{blacklist} class, it does not affect our results as it was not used in the experiments.

Additionally, a high portion of the \emph{anomaly-sshscan} class flows was omitted from the test set (over 60\%). However, as displayed in Table~\ref{tab:01}, such a class was not included in the test set at all. This phenomenon means it had to be outvoted while determining aggregated sample class labels. To clarify the previous statement, let us explain the process of sample label assignment during the preprocessing phase: A situation when multiple flows with different class labels (e.g., some \emph{background}, some with a specific type of attack) are aggregated into the same 3-minute window sample, may happen. In such a case, the class label included in more than half of aggregated flows is used (i.e., the majority voting).

\begin{table}[b]
\caption{The Omitted and Outvoted Flow Labels in the Train and Test Sets}
\begin{center}
\begin{tabular}{|c|c|c||c|c|}
\hline
 & \multicolumn{2}{c||}{\textbf{Omitted}} & \multicolumn{2}{c|}{\textbf{Outvoted}} \\
\cline{2-5}
\textbf{Class label} & \textbf{Train} & \textbf{Test} & \textbf{Train} & \textbf{Test} \\
\hline
background & 20.76\% & 21.24\% & 0.01\% & 0.02\% \\
blacklist & 14.60\% & 18.79\% & 45.15\% & 40.08\% \\
nerisbotnet & 0.26\% & 0.24\% & 6.72\% & 7.27\% \\
anomaly-spam & 7.28\% & 0.34\% & 8.16\% & 0.26\% \\
dos & 0.00\% & 0.00\% & 1.09\% & 0.00\% \\
scan44 & 0.00\% & 0.005\% & 0.00\% & 0.00\% \\
scan11 & 0.00\% & 0.00\% & 0.00\% & 0.04\% \\
anomaly-udpscan & 0.00\% & 0.00\% & 0.00\% & 0.00\% \\
anomaly-sshscan & 0.00\% & 61.11\% & 75.00\% & 85.71\% \\
\hline
\end{tabular}
\label{tab:06}
\end{center}
\vspace{-5mm}
\end{table}

The problem of outvoting is showcased in the right part of Table~\ref{tab:06} for each class label. These results were obtained by calculating the percentage of flows with minority class labels during the aggregation process with respect to all aggregated flows. As indicated, outvoting affected mostly the \emph{blacklist} and \emph{anomaly-sshscan} classes. Since neither of them was used in our evaluation, the experimental results were not influenced by this issue. The \emph{anomaly-udpscan} class contained no flow records in the selected subset of the UGR'16 dataset. Therefore, it has a 0.00\% value in all cases (omitted flow records as well as outvoted flow records), so it was not provided in Table~\ref{tab:01}.

\subsection{Temporal Experimental Bias}
\label{ssec:discussion_tempbias}

As briefly outlined back in Section~\ref{ssec:04_ugr_dataset}, we are aware of a potential temporal bias because all objects in the training data are not strictly temporally precedent to the testing data~\cite{pendlebury2019_tesseract}). Such a temporal bias might skew the evaluation and produce over-optimistic results not corresponding to reality, as elaborated on by Pendlebury et al.~\cite{pendlebury2019_tesseract}.

\begin{figure}[b]
    \vspace{-1.25em}
    \centering
    \includegraphics[width=.95\linewidth,height=6cm,keepaspectratio]{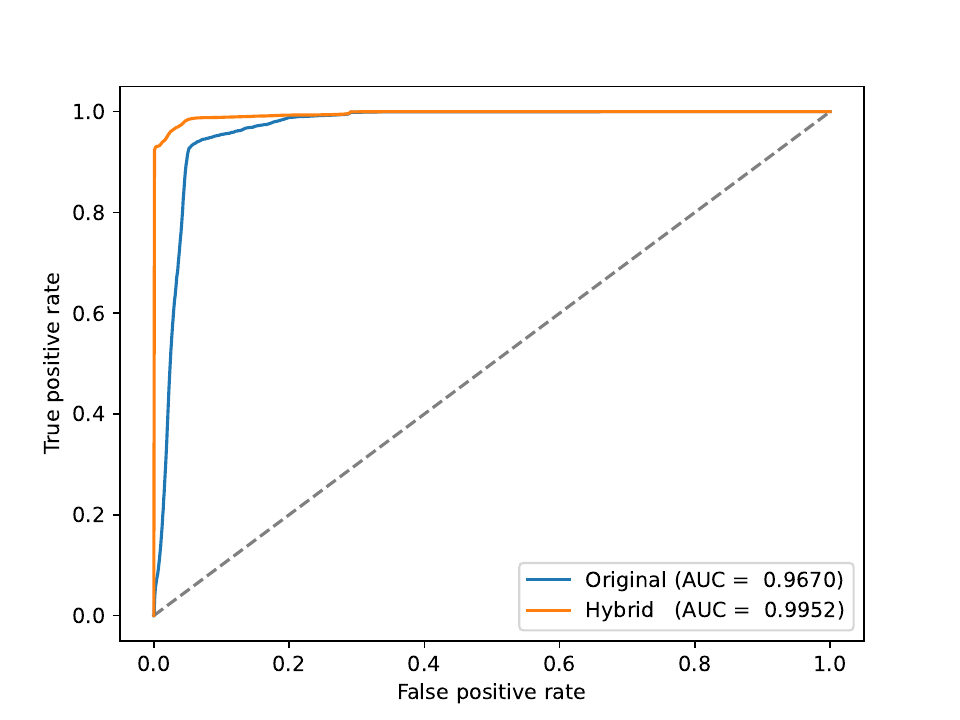}
    \caption{Comparison of ROC curves and AUC values of the original and hybrid models without temporal bias.}
    \label{fig:auc_kde_notempbias}
    \vspace{-5mm}
\end{figure}

Nevertheless, all experiments performed so far have used this data to retain comparability with the original GEE study~\cite{nguyen2019gee}, which used the same subset. Nevertheless, in order to measure the potential impact of temporal bias, we selected another day -- August 8th, as the new test dataset subset. We then reran the hybrid detection pipeline to determine the AUC metric of the unsupervised anomaly detector with and without a supervised element (\figurename~\ref{fig:auc_kde_notempbias}).

August 8 is strictly preceded by the training days (March 19, July 30) and thus retains the data temporal consistency. We decided to use this day because it was the only day that contained the same attack classes as the train subset. After the data processing using time-windows, we determine that the subset contains $1.09$M \emph{background} samples, around $5,800$ \emph{anomaly-spam} samples, and less than $100$ for \emph{dos}, \emph{nerisbotnet}, and \emph{scan} classes each.

Interestingly, the original unsupervised anomaly detector performed even better than in the previous subset, achieving an AUC value of $0.97$. The hybrid approach achieved an AUC of $0.9952$, similar to $0.9977$ for the previous test set. This fact demonstrates that the results were not significantly skewed by a potential temporal bias, and the approach performs well regardless. Also, it is much more stable than the original detector (minor change in AUC).

This set of experiments does not indeed imply that there was no potential temporal bias in the previous test set present. Also, it does not imply that the employed method is resistant to temporal bias. The point to show here is that the findings presented throughout this paper (especially focusing on a supervised prefilter impact) are relevant despite this fact.

As the August 8 (test day) is rather close to the July 30 (training day), a more severe drift may not have been present yet. Nevertheless, to keep the performance high, practical implementations would require constant retraining of both supervised and unsupervised modules to adapt to new traffic profiles and emerging attacks. For such cases, the paradigm of active learning~\cite{settles2009_active_learning_survey} could be used to label some portion of the traffic (typically with the lowest model's confidence) and use it to update the model. One such promising approach was presented by Andersini et al., named INSOMNIA~\cite{andersini2021_insomnia}.

\subsection{Comparison With Other Approaches}
\label{ssec:discussion_comparison}

As mentioned in Section~\ref{sec:02}, although various studies on hybrid NIDSs have been published, analyses of the supervised element's impact on anomaly detection are scarce. In contrast to similar work by Kim et al.~\cite{kim2014novel}, we analyze the impact on autoencoder-based anomaly detection, which is much more popular currently~\cite{yang2022_slr_anids,habeeb2022_nids_survey,ahmad2021_nids_systematic_study_ml}. In addition, we perform our experiments using a more realistic and newer dataset (i.e., containing real background ISP traffic). As a result, we expect our results to provide better insights into how the method (and the impact of the supervised prefilter on anomaly detection) might work in real-world scenarios.

\begin{table}[b]
\centering
\small
\caption{Comparison of the employed hybrid approach to other methods. Abbreviations explanation: DR -- Detection Rate, FA -- False Alarm rate, F1 -- F1-score.}
\begin{tabular}{|m{2.2cm}|m{2.5cm}|m{1cm}|m{1.25cm}|}
\hline
\textbf{Paper} & \textbf{Method} & \textbf{Dataset} & \textbf{Results} \\ \hline
     Zhang et al.~\cite{zhang2008random} &  2 Random Forests & KDD'99 & \makecell[l]{DR: $0.944$ \\ FA: $0.011$} \\ \hline
     Elbasiony et al.~\cite{elbasiony2013_hybrid_nids} &  Random Forest + K-means & KDD'99 & \makecell[l]{DR: $0.983$ \\ FA: $0.016$} \\ \hline
     Kim et al.~\cite{kim2014novel} &  C4.5 Decision Tree + 1-class SVM & NSL-KDD & \makecell[l]{Known:\\DR: $0.991$ \\ FA: $0.012$\\ Unknown:\\ DR: $0.875$ \\ FA: $0.050$ } \\ \hline
     Bangui et al.~\cite{bangui2022hybrid} &  Random Forest + Coresets & CIC-IDS2017 & \makecell[l]{F1: $0.944$} \\ \hline
     This Work &  Random Forest + Variatonal Autoencoder &  UGR'16 & \makecell[l]{DR: $0.998$\\ FA: $0.132$\\F1: $0.929$} \\ \hline
\end{tabular}
\label{tab:methods_comparison}
\end{table}

Although not the main objective of this paper, we also tried to compare our hybrid detector to other proposals (Table~\ref{tab:methods_comparison}). For this purpose, we have focused on the type 2 Hybrid NIDSs (misuse module first) using a random forest or other tree-based methods. However, due to the dataset choice and a specific subset within it, the results cannot be directly compared (except for the original GEE paper using the same dataset and its subset). Nevertheless, we are still including a table for a reader to understand the approximate performance expected from hybrid methods deployed in the wild.

\section{Conclusion}
\label{sec:conclusion}

As discussed throughout this paper, supervised learning approaches offer a decent detection rate of learned anomaly patterns but suffer when dealing with new ones. For this reason, they are often combined with unsupervised anomaly detection into hybrid approaches. Such solutions are generally expected to offer better results.

Although various hybrid approaches have been explored previously, not many papers focused on analyzing methods' synergy and measuring their mutual impact. Therefore, we aimed to address this knowledge gap by extending a state-of-the-art anomaly detection method into a hybrid approach. Firstly, all traffic is processed by a supervised binary classifier. The traffic classified as benign is then processed by an original anomaly detector -- a variational autoencoder to detect potential new attacks.

Our experiments with the UGR'16 dataset have shown that the detector's AUC increased by over 11\%, detecting 30\% more attacks than the original unsupervised state-of-the-art anomaly detector. This finding implies that anomaly detectors might not be able to catch all the attacks because they do not cause any anomalous behavior. Therefore, employing supervised classifiers is reasonable not only for reducing false positives but also for increasing the detection rate. To highlight this fact, we have intentionally set a relatively low reconstruction error threshold, leading to approximately the same number of false positives. The threshold can indeed be increased to achieve a lower false alarm rate at the cost of a worsened attack detection rate. Nevertheless, the results imply that supervised learning improves the sole unsupervised approach but is still inferior to a hybrid method with both supervised and unsupervised elements. This fact was demonstrated via zero-day attack simulation using novelty tests.

In future work, we aim to experiment with tweaking time window length to improve anomaly detection latency. An optimal window length retains most of the detection capabilities while performing the detection as fast as possible. Depending on a potential deployment environment, more work could be dedicated to exploring multi-class classifiers to obtain attack explainability while retaining as high classification performance as possible. This could be achieved via hyperparameter tuning of both supervised and unsupervised parts. Last but not least, the model should withstand gradual changes in network characteristics (concept drift). Therefore, a mechanism that would automatically adapt to new patterns without significant manual intervention would be very beneficial.

\bibliographystyle{ACM-Reference-Format}
\bibliography{sources}

\end{document}